\documentclass[twoside,11pt]{article}
\usepackage{jmlr2e}
\usepackage{amsmath}

\usepackage{listings}
\usepackage{caption}
\newcommand{\comment}[1]{}

\newcommand{\jtv}[1]{{\em JTV: #1}}
\newcommand{\bit}{\begin{itemize}}
\newcommand{\eit}{\end{itemize}}

\newcommand{\mmani}{{\tt megaman}}
\newcommand{\sklearn}{{\tt scikit-learn}}

\newcommand{\bigOO}{{\cal O}}

\ShortHeadings{megaman: Manifold Learning with Millions of points}{McQueen, Meil\u{a}, VanderPlas and Zhang}
\firstpageno{1}

\title{megaman: Manifold Learning with Millions of points}

\author{\name James McQueen \email jmcq@uw.edu \\
       \addr Department of Statistics\\
       University of Washington\\
       Seattle, WA 98195-4322, USA
       \AND
       \name Marina Meil\u{a} \email mmp@stat.washington.edu \\
       \addr Department of Statistics\\
       University of Washington\\
       Seattle, WA 98195-4322, USA
       \AND
       \name Jacob VanderPlas \email jakevdp@uw.edu\\
       \addr e-Science Institute\\
        University of Washington\\
       Seattle, WA 98195-4322, USA
       \AND
       \name Zhongyue Zhang \email zhangz6@cs.washington.edu\\
       \addr Department of Computer Science and Engineering\\
        University of Washington\\
       Seattle, WA 98195-4322, USA
}

\editor{}
\begin{document}
\maketitle

\begin{abstract}%<- trailing % for backward compatibility of .sty file
	\emph{Manifold Learning (ML)} is a class of algorithms seeking a low-dimensional
    non-linear representation of high-dimensional data.
    Thus ML algorithms are, at least in theory, most applicable to high-dimensional
    data and sample sizes to enable accurate estimation of the manifold.
    Despite this, most existing manifold learning implementations are not particularly scalable.
    Here we present a Python package that implements a variety of manifold learning algorithms
    in a modular and scalable fashion, using fast approximate neighbors searches and fast
    sparse eigendecompositions. The package incorporates theoretical advances in manifold learning, such as the unbiased Laplacian estimator introduced by \cite{coifman:06} and the estimation of the embedding distortion by the Riemannian metric method introduced by \cite{2013arXiv1305.7255P}. 
    In benchmarks, even on a single-core desktop computer, our code embeds millions of data points
    in minutes, and takes just 200 minutes to embed the main sample of galaxy spectra from the
    Sloan Digital Sky Survey --- consisting of 0.6 million samples in 3750-dimensions  ---
    a task which has not previously been possible.
\end{abstract}


\comment{it has been shown statistically that the estimation accuracy
 depends asymptotically on the sample size $N$ like $N^{1/(\alpha d +\beta)}$,
 hence requires large amounts of data when the intrinsic dimension $d$ is larger than 1 or 2.
 Moreover dimension reduction is more beneficial when the input data is high-dimensional.}

\comment{fully realizes its potential in
           scientific discovery from very high dimensional data that
           can be described by a small number of parameters
           from very large multi-dimensional data
           sets representing partially known physical systems,
           (e.g. spectra of galaxies) where there is reason to believe
           that the data can be modeled by a small set of parameters.}

\section{Motivation}

\emph{Manifold Learning (ML)} algorithms like Diffusion Maps or Isomap
find a non-linear representation of high-dimensional data with a small
number of dimensions. Research in ML is making steady progress, yet
there is currently no ML software library efficient enough to be used
in realistic research and application of ML.

The first comprehensive attempt to bring several manifold learning
algorithms under the same umbrella is the {\tt mani} Matlab package
authored by \cite{wittman:10}. This package was instrumental in
illustrating the behavior of the existing ML algorithms of a variety
of synthetic toy data sets. More recently, a new Matlab toolbox {\tt
  drtoolbox}\footnote{{\tt https://lvdmaaten.github.io/drtoolbox/}}
was released, that implements over thirty dimension reduction
methods, and works well with sample sizes in the thousands.

Perhaps the best known open implementation of common manifold learning
algorithms is the {\tt manifold} sub-module of the Python package
\sklearn\footnote{{\tt
    http://scikit-learn.org/stable/modules/manifold.html}}
\citep{pedregosa2011}. This software benefits from the integration with \sklearn, meets its standards and
philosophy, comes with excellent documentation and examples, and is
written in a widely supported open language.

The \sklearn{} package strives primarily for usability rather than scalability.
While the package does feature some algorithms designed with scalability in
mind, the manifold methods are not among them.
For example, in \sklearn{} it is difficult for
different methods to share intermediate results, such as eigenvector
computations, which can lead to inefficiency in data exploration. The
\sklearn{} manifold methods cannot handle out-of-core data, which
leads to difficulties in scaling. Moreover, though \sklearn{} accepts
sparse inputs and uses sparse data structures internally, the current
implementation does not always fully exploit the data sparsity.

To address these challenges, we propose \mmani{}, a new Python package for
scalable manifold learning.
This package is designed for performance, while adopting the look and feel
of the \sklearn{} package, and inheriting the functionality of
\sklearn's well-designed API \citep{buitinck2013}.

\section{Background on non-linear dimension reduction and manifold learning}

This section provides a brief description of non-linear dimension
reduction via manifold learning, outlining the tasks performed by a
generic manifold learning algorithm. The reader can find more
information on this topic in \cite{2013arXiv1305.7255P}, as well as on  the
\sklearn{} web site\footnote{{\tt http://scikit-learn.org/stable/auto\_examples/index.html}}.

We assume that a set of $N$ vectors $x_1,\ldots x_N$ in $D$ dimensions
is given (for instance, for the data in Figure \ref{fig:word2vec}, $N=3\times 10^{6},\,D=300$). It is assumed that these data are sampled from (or
approximately from) a lower dimensional space (i.e. a {\em manifold})
of {\em intrinsic dimension} $d$. The goal of manifold learning is to
map the vectors $x_1,\ldots x_N$ to $s$-dimensional vectors
$y_1,\ldots y_N$, where $y_i$ is called the \emph{embedding} of $x_i$
and $s\ll D,\,s\geq d$ is called the \emph{embedding dimension}. The
mapping should preserve the neighborhood relations of the original
data points, and as much as possible of their local geometric
relations.

From the point of view of a ML algorithm (also called an
\emph{embedding algorithm}), non-linear dimension reduction subsumes
the following stages.

Constructing a \emph{neighborhood graph} $G$, that connects all point
pairs $x_i,x_j$ which are ``neighbors''. The graph $G$ is sparse when
data can be represented in low dimensions.  From this graph, a
similarity $S_{ij}\geq 0$ between any pair of points $x_i,x_j$ is
computed by $S_{ij}=\exp\left(-||x_i-x_j||^2/\sigma^2\right)$, where
$\sigma$ is a user-defined parameter controlling the neighborhood
size. This leads to a $N\times N$ \emph{similarity matrix} $S$,
sparse. Constructing the neighborhood graph is common to most existing
ML algorithms.

From $S$ a special $N\times N$ symmetric matrix called the
\emph{Laplacian} is derived. The Laplacian is directly used for
 embedding in  methods such as the Spectral Embedding
(also known as Laplacian Eigenmaps) of \cite{belkin:01} and Diffusion
Maps of \cite{nadler:06}. The Laplacian is also used after embedding,
in the post-processing stage.  Because these operations are common to
many ML algorithms and capture the geometry of the high-dimensional
data in the matrices $S$ or $L$, we will generically call the software
modules that implement them \emph{Geometry}.

The next stage, the \emph{Embedding} proper, can be performed by
different {\em embedding algorithms}. An embedding algorithm take as
input a matrix, which can be either the matrix of distances
$||x_i-x_j||$ or a matrix derived from it, such as $S$ or $L$. Well
known algorithms implemented by \mmani{} are Laplacian Eigenmaps
\citep{belkin:01}, Diffusion Maps \citep{nadler:06}, Isomap
\citep{bernsteinDeSilvaLangfordTenn:00}, etc. The ouput of this step
is the $s$-dimensional representation $y_i$ corresponding to each data
point $x_i$. Embedding typically involves computing $\bigOO(d)$
eigenvectors of some $N\times N$ matrix. To note that this matrix has
the same sparsity pattern as the neighborhood graph $G$.
\begin{figure}
\centerline{\includegraphics[]{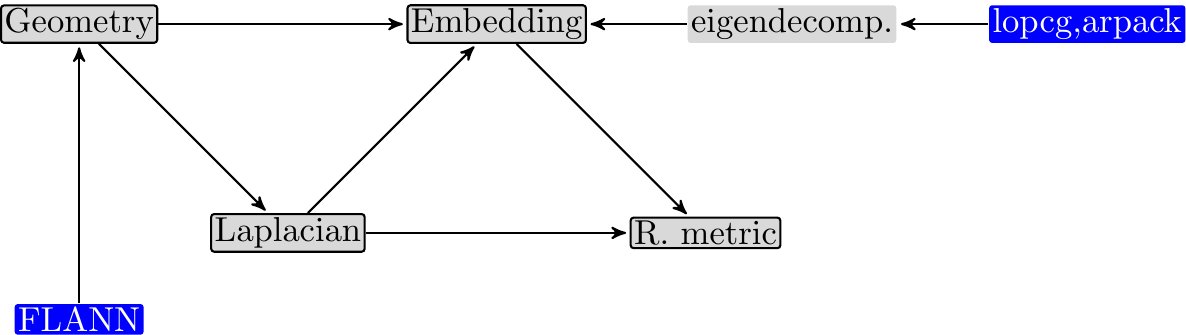}}
	\caption{\label{fig:ml-tasks}
\mmani{} classes (gray, framed), packages (gray, no frame) and external packages (blue). The class structure reflects the relationships between ML tasks.
}

\end{figure}
Finally, for a given embedding $y_1,\ldots y_n$, one may wish to
estimate the distortion incurred w.r.t the original data. This can be
done via the method of \cite{2013arXiv1305.7255P}, by calculating for
each point $y_i$ a \emph{metric}\footnote{ Mathematically speaking
  $R_i$ represents the inverse of the push-forward Riemannian metric in
  the embedding space \citep{2013arXiv1305.7255P}.} $R_i$; $R_i$ is an
$s\times s$ symmetric, positive definite matrix. Practically, the
value $u^TR_iu$, with $u$ a unit vector, is the ``stretch'' at $y_i$ in
direction $u$ of the embedding $\{y_1,\ldots, y_N\}$ w.r.t the original
data $\{x_1,\ldots x_N\}$. In particular, for an embedding with no
stretch\footnote{That is, for an \emph{isometric} embedding.}, $R_i$
should be equal to the unit matrix. The metric given by $R_1,\ldots R_N$ can be used to calculate distances, areas and volumes using the embedded data, which approximate well their respective values on the original high-dimensional data. Therefore, obtaining $R_{1:N}$ along with the embedding coordinates $y_{1:N}$ provides a geometry-preserving (lossy) compression of the original $x_{1:N}$.

The logical structure of these tasks is shown in Figure
\ref{fig:ml-tasks}, along with some of the classes and software
packages used to implement them.

\begin{figure}
\centering \includegraphics[scale =
  0.7]{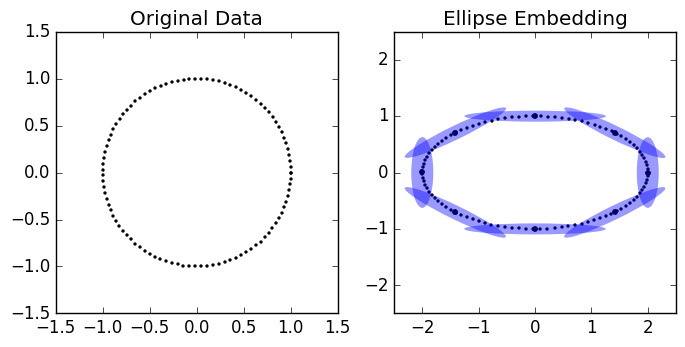} \captionof{figure}{
  \footnotesize The circle on the left hand plot represents the
  original data with $D=2$.  The ellipse in the right hand plot is an
  ``embedding'' of these data into $s=2$ dimensions, which deforms the
  circle into an ellipse. The blue ovals plotted at selected points
  represent the (dual) Riemannian Metric at that point. The long axes
  of the ellipses represent the local unit of length along the ellipse
  \comment{the short axes are artifacts}; the units are smaller left
  and right, where the length of the original curve was compressed,
  and larger in the middle where the circle was stretched. }
\label{fig:rmetric}
\end{figure}

\section{Software design}
\noindent
\subsection{Functionality}
We provide classes that implement the above tasks, along with
post-processing and visualization tools.
The package implements recent advances in the statistical
understanding of manifold learning such as:

\bit
 \item It has been shown decisively by \cite{HeinAL:07} and \cite{TingHJ:10} that
 		the construction of the neighborhood graph can influence the resulting
 		embedding. In particular, the $k$-nearest neighbor graph construction
 		introduces biases that can be avoided by the use of the $\epsilon$-radius
 		graphs. Therefore, the default graph construction method is {\tt radius\_neighbors}.
\item A variety of graph Laplacians
		({\tt unnormalized, normalized, random\_walk} \citep{vonluxburg:07},
		{\tt renormalized} and {\tt geometric} \citep{coifman:06}) are available. The default choice is the
		{\tt geometric} Laplacian introduced by \citep{coifman:06} who showed
		that this choice eliminates the biases due variation in the density
		of the samples.\footnote{Technically speaking, this is a type of
		renormalized Laplacian, which converges to the Laplace-Beltrami operator
		if the data is sampled from a manifold.}
\item  Embedding metric estimation following \cite{2013arXiv1305.7255P}.  The (dual)
		Riemannian metric measures the ``stretch'' introduced by the
		embedding algorithm with respect to the original data; Figure
		\ref{fig:rmetric} illustrates this.
\eit
\noindent

\subsection{Designed for performance}

The \mmani{} package is designed to be simple to use with an interface
that is similar to the popular \sklearn{} python package.
Additionally, \mmani{} is designed with experimental and exploratory
research in mind.

\subsubsection{Computational challenges}

The ML pipeline described above and in Figure \ref{fig:ml-tasks}
comprises two significant computational bottlenecks.

{\bf Computing the neighborhood graph} involves finding the
$K$-nearest neighbors, or the $r$-nearest neighbors (i.e all neighbors
within radius $r$) in $D$ dimensions, for $N$ data points. This leads
to sparse graph $G$ and sparse $N\times N$ matrix $S$ with $\bigOO(d)$
neighbors per node, respectively entries per row.  All $N \times N$ matrices in subsequent steps
will have the same sparsity pattern as $S$. Naively
implemented, this computation requires $\bigOO(N^2(D+\log N))$
operations. 

{\bf Eigendecomposition} A large number of embedding algorithms
(effectively all algorithms implemented in \mmani) involve computing
$\bigOO(s)$ principal eigenvectors of an $N\times N$ symmetric,
semi-positive definite matrix. This computation scales like
$\bigOO(N^2s)$ for dense matrices.
In addition, if the working matrices were stored in dense form, the
memory requirements would scale like $N^2$, and elementary linear
algebra operations like matrix-vector multiplications would scale
likewise.

\subsubsection{Main design features}
We made several design and implementation choices in support of scalability.
\bit
	\item Sparse representation are used as default for optimal
		storage and computations.
	\item We incorporated state of the art
		\underline{F}ast \underline{L}ibrary
		\underline{A}pproximate \underline{N}earest
		\underline{N}eighbor search algorithm by \cite{flann:2014}
		(can handle Billions of data points) with a cython
		interface to Python allowing for rapid neighbors
		computation.
	\item  Intermediate data structures (such as data set index,
		distances, affinity matrices, Laplacian) are cached allowing for
		testing alternate parameters and methods without redundant
		computations.
	\item By converting matrices to sparse symmetric positive definite
		(SSPD) in all cases, \mmani{} takes advantage of {\tt pyamg} and of the
		\underline{L}ocally \underline{O}ptimal \underline{B}lock
		\underline{P}reconditioned \underline{C}onjugate \underline{G}radient
		(LOBPCG) package as a matrix-free method \citep{lobpcg:2001} for solving generalized eigenvalue
		problem for SSPD matrices. Converting to symmetric matrices also improves the numerical stability of the algorithms.
\comment{	\item Lazy evaluation in post-processing
	 	(specifically the {\tt rmetric}/''stretch'' evaluation)}
\eit
\noindent
With these, \mmani{} runs in reasonable time on data sets in the
millions.  The design also supports well the exploratory type of work
in which a user experiments with different parameters (e.g.  radius or
bandwidth parameters) as well as different embedding procedures;
\mmani{} caches any re-usable information in both the {\tt Geometry}
and {\tt embedding} classes.

\comment{Redundant? place it somewhere earlier?
It is well-known that manifold learning methods require
large data sets to perform well. Unfortunately, many of these
methods appear computationally unfeasible with large data sets.
\mmani{} attempts to overcome these technical challenges by harnessing
existing tools for solving the intermediate problems and unifying
them under a single framework. The FLANN C++ package is used
(via cython) for the pairwise neighbors calculation. Additionally
sparse representations are used to minimize storage requirements
and increase computing various matrices. Finally by manipulating
the Laplacian and other embedding matrices we can convert them
to SSPD in order to take
advantage of the matrix-free LOBPCG procedure to calculate the
eigendecomposition.}

\subsection{Designed for extensions}
\noindent
\mmani's interface is similar to that of the \sklearn{} package,
in order to facilitation easy transition for the users of \sklearn.

\mmani{} is object-oriented and modular. For example, the {\tt
  Geometry} class provides user access to geometric computational
procedures (e.g. fast approximate radius neighbors, sparse affinity
and Laplacian construction) as an independent module, whether the user
intends to use {\tt embedding} methods or not.  \mmani{} also offers
the unified interface {\tt eigendecomposition} to a handful of
different eigendecomposition procedures (dense, {\tt arpack}, {\tt
  lobpcg}, {\tt amg}). Consequently, the \mmani{} package can be used
for access to these (fast) tools without using the other classes. For
example, \mmani{} methods can be used to perform the Laplacian
computation and embedding steps of spectral clustering. Finally, {\tt
  Geometry} accepts input in a variety of forms, from data cloud to
similarity matrix, allowing a user to optionally input a precomputed
similarity.

This design facilitates easy extension of \mmani{}'s functionality.
In particular, new embedding algorithms and new methods for distance
computation can be added seamlessly. More ambitious possible
extensions are: neighborhood size estimation, dimension estimation,
Gaussian process regression\comment{integrated K-nearest neighbors (currently
only radius neighbors natively supported but a pre-computed K-nearest
neighbors matrix can be passed)}.

Finally, the \mmani{} package also has a comprehensive documentation with
examples and unit tests that can be run with nosetests to ensure validity.
To allow future extensions \mmani{} also uses Travis Continuous 
Integration\footnote{https://travis-ci.org/}.

\section{Downloading and installation}
\mmani{} is publically available at: {\tt https://github.com/mmp2/megaman}.
\mmani's required dependencies are {\tt numpy}, {\tt scipy},
and {\tt scikit-learn}, but for optimal performance FLANN,
{\tt cython},  {\tt pyamg} and a c compiler {\tt gcc} are also required.
For unit tests and integration \mmani{} depends on {\tt nose}.
The most recent \mmani{} release can be installed along with its dependencies using the cross-platform conda\footnote{http://conda.pydata.org/miniconda.html} package manager:
{\tt
\begin{lstlisting}
$ conda install -c https://conda.anaconda.org/jakevdp megaman
\end{lstlisting}
}
\noindent
Alternatively, the \mmani{} can be installed from source by downloading the source repository and running:
{\tt
\begin{lstlisting}
$ python setup.py install
\end{lstlisting}
}
\noindent
With {\tt nosetests} installed, unit tests can be run with:
{\tt
\begin{lstlisting}
$ make test
\end{lstlisting}
}

\section{Quick start}
For full documentation see the \mmani{} website at: {\tt http://mmp2.github.io/megaman/}
{\tt
\begin{lstlisting}
from megaman.geometry import Geometry
from megaman.embedding import SpectralEmbedding
from sklearn.datasets import make_swiss_roll

n = 10000
X, t = make_swiss_roll( n )
n_components = 2
radius = 1.1
geom = Geometry(adjacency_method = 'cyflann',
		adjacency_kwds = {'radius':radius},
		affinity_method = 'gaussian',
		affinity_kwds = {'radius':radius},
		laplacian_method = 'geometric',
		laplacian_kwds = {'scaling_epps':radius})
SE = SpectralEmbedding(n_components=n_components,
                       eigen_solver='amg',
                       geom=geom)
embed_spectral = SE.fit_transform(X)
\end{lstlisting}
}% end tt

\section{Classes overview}
The following is an overview of the classes and other
tools provided in the \mmani{} package:

	{\tt Geometry} This is the primary non-embedding class of
		the package. It contains functions to compute the
		pairwise distance matrix (interfacing with the distance module),
		the Gaussian kernel similarity matrix $S$ and the Laplacian
		matrix $L$. A {\tt Geometry} object is what is passed or created
		inside the embedding classes.

	{\tt RiemannianMetric} This class produces the estimated Riemannian
		metric $R_i$ at each point $i$ given an embedding $y_{1:N}$
		and the estimated Laplacian from the original data.

	{\tt embeddings} The manifold learning algorithms are implemented
		in their own classes inheriting from a base class. Included are:
		\bit
			\item {\tt SpectralEmbedding} implements \emph{Laplacian Eigenmaps} \citep{belkin:01} and \emph{Diffusion Maps} \citep{nadler:06}:
				These methods use the eigendecomposition of the
				Laplacian.
			\item {\tt LTSA} implements the \emph{Local Tangent Space Alignment} method \citep{ZhangZ:04}: This method
				aligns estimates of local tangent spaces.
			\item  {\tt LocallyLinearEmbedding} (LLE)
                          \citep{roweis:00}: This method finds
                          embeddings that preserve local
                          reconstruction weights
			\item {\tt Isomap} \citep{bernsteinDeSilvaLangfordTenn:00}:
				This method uses Multidimensional Scaling to preserve shortest-path
                distances along a neighborhood-based graph.

		\eit
{\tt eigendecomposition} Not implemented as a class, this  module
		provides a unified (function) interface to the different
		eigendecomposition methods provided in {\tt scipy}. It also
		provides a null space function (used by LLE and LTSA).

\section{Benchmarks}
\begin{figure}
%\centering
\hspace{-2em}
\begin{minipage}{.45\textwidth}
 % \centering
  \includegraphics[scale = 0.4]{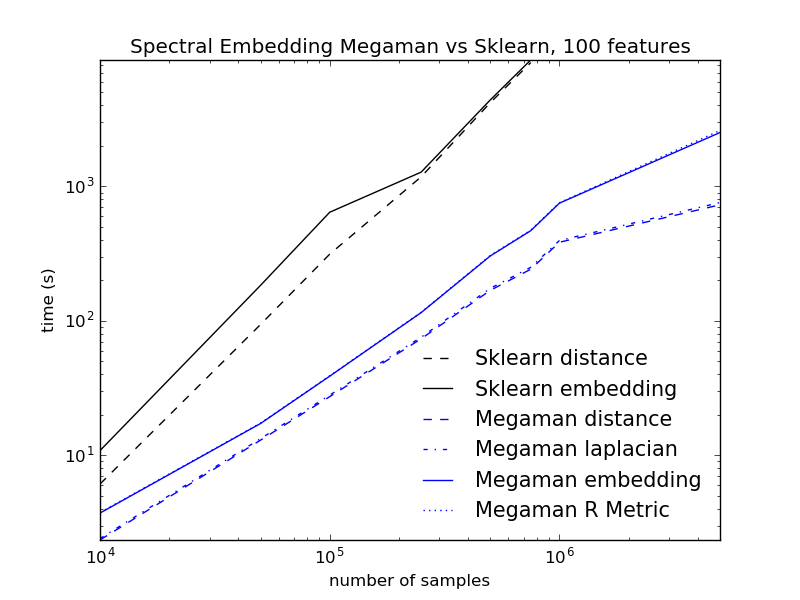}
  \captionof{figure}{
  			\footnotesize
		  {\bf Run time vs. data set size $N$} for
          fixed $D=100$ and $\epsilon=\frac{5}{N^{1/8}}+ D^{1/4} - 2$.
          The data is from a Swiss Roll
          (in 3 dimensions) with an additional 97 noise dimensions,
          embedded into $s=2$ dimensions by the {\tt spectral\_embedding} algorithm.
          By $N = 1,000,000$ \sklearn{} was unable to compute an
          embedding due to insufficient memory. All \mmani{} run times (including time between 
          distance and embedding) are faster than \sklearn{}.}
  \label{fig:bench-n}
\end{minipage}%
\hfill
\begin{minipage}{.45\textwidth}
%  \centering
\hspace{-1em}
  \includegraphics[scale = 0.4]{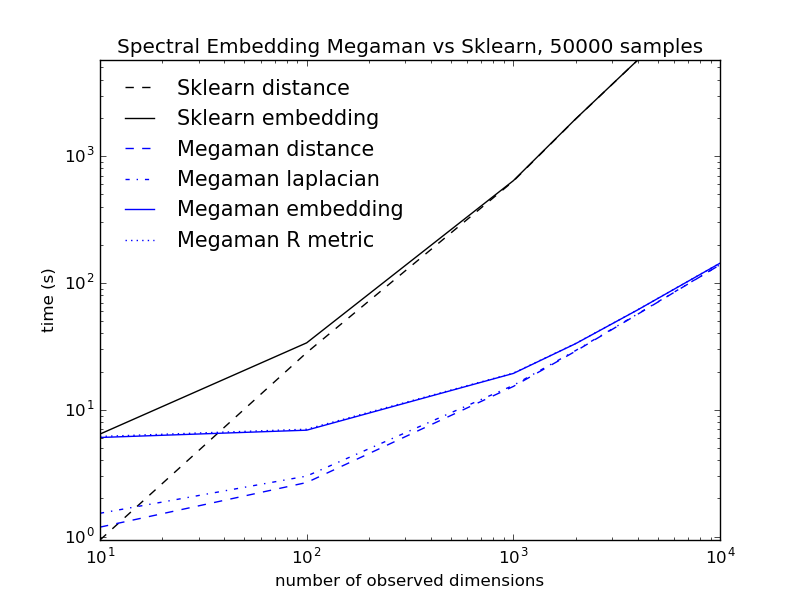}
  \captionof{figure}{\footnotesize
  			{\bf Run time vs. data set dimension $D$} for
            fixed $N=50,000$ and $\epsilon=\frac{5}{N^{1/8}} + D^{1/4} - 2$.
            The data is from a Swiss Roll (in 3 dimensions)
            with additional noise dimensions, embedded into $s=2$
            dimensions by the {\tt spectral\_embedding} algorithm.
            By $D = 10,000$ \sklearn{} was unable to compute an
            embedding due to insufficient memory. All \mmani{} run times (including time between 
          distance and embedding) are faster than \sklearn{}.}  \label{fig:bench-D}
\end{minipage}
\end{figure}

The one other popular comparable implementation of manifold learning
algorithms is the \sklearn{} package.

To make the comparison as fair as possible, we choose the Spectral
Embedding method for the comparison, because both \sklearn{} and \mmani{}
allow for similar settings:
\bit
  \item Both are able to use radius-based neighborhoods
  \item Both are able to use the same fast eigensolver, a {\it Locally-Optimized Block-Preconditioned Conjugate Gradient (lobpcg)}
        using the algebraic multigrid solvers in {\it pyamg}.
\eit
Incidentally, Spectral Embedding is empirically the fastest
of the usual manifold learning methods, and the best understood
theoretically. Note that with
the default settings, \sklearn~ would perform slower than in our
experiments.

We display total embedding time (including time to compute the graph
$G$, the Laplacian matrix and the embedding) for \mmani{} versus
\sklearn, as the number of samples $N$ varies (Figure
\ref{fig:bench-n}) or the data dimension $D$ varies (Figure
\ref{fig:bench-D}).  All benchmark computations were performed on a
single desktop computer running Linux with 24.68GB RAM and a Quad-Core
3.07GHz Intel Xeon CPU. We use a relatively weak machine to
demonstrate that our package can be reasonably used without high
performance hardware. 

The experiments show that \mmani{} scales considerably better than
\sklearn, even in the most favorable conditions for the latter; the
memory footprint of \mmani{} is smaller, even when \sklearn~ uses
sparse matrices internally. The advantages grow as the data size
grows, whether it is w.r.t $D$ or to $N$.

The two earlier identified bottlenecks, distance computation and
eigendecomposition, dominate the compute time. By comparison, the
other steps of the pipe-line, such as Laplacian and Riemannian metric
computation are negligible. When $D$ is large, the distance
computation dominates, while for $N\ll D$, the eigendecomposition
takes a time comparatively equal to the distance computation.

\comment{
The primary bottlenecks to the embedding procedures are the
neighborhood graph construction and the eigendecomposition.
In order to optimize the speed of the package on large
data use option `distance\_method' $=$ 'cyflann' should be chosen
which directly interfaces with the FLANN C++ package to compute
the neighborhood graph. The fastest method for computing
the eigendecomposition is with LOBPCG using preconditioning
from the pyamg package, option `eigen\_decomp' $=$ 'pyamg'. Finally,
not all embedding method scale equally. The fastest method
for use on largest scale data is the Spectral/Diffusion Maps.
Locally Tangent Space alignment scales better than LLE and
Isomap itself is fundamentally non-scalable as it requires
the eigendecomposition of an \emph{dense} $N \times N$ matrix
of graph shortest paths.
}

\begin{figure}
\centering
\hfill
\begin{minipage}{.45\textwidth}
  \centering
  \includegraphics[scale = 0.52]{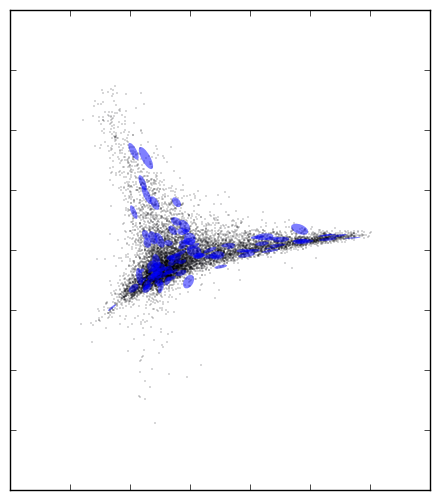}
  \captionof{figure}{\footnotesize
{3,000,000 words and phrases mapped by {\tt word2vec} into 300
  dimensions were embedded into 2 dimensions using Spectral
  Embedding. The plot shows a sample of 10,000 points displaying the overall
  shape of the embedding as well as the estimated ``stretch'' (i.e. dual push-forward
  Riemannian metric) at various locations in the embedding. \comment{The
    figure above shows a sample of points and a few example words
    highlighted.}}}
  \label{fig:word2vec}
\end{minipage}
\hfill
\begin{minipage}{.45\textwidth}
  \centering
  \includegraphics[scale = 0.25]{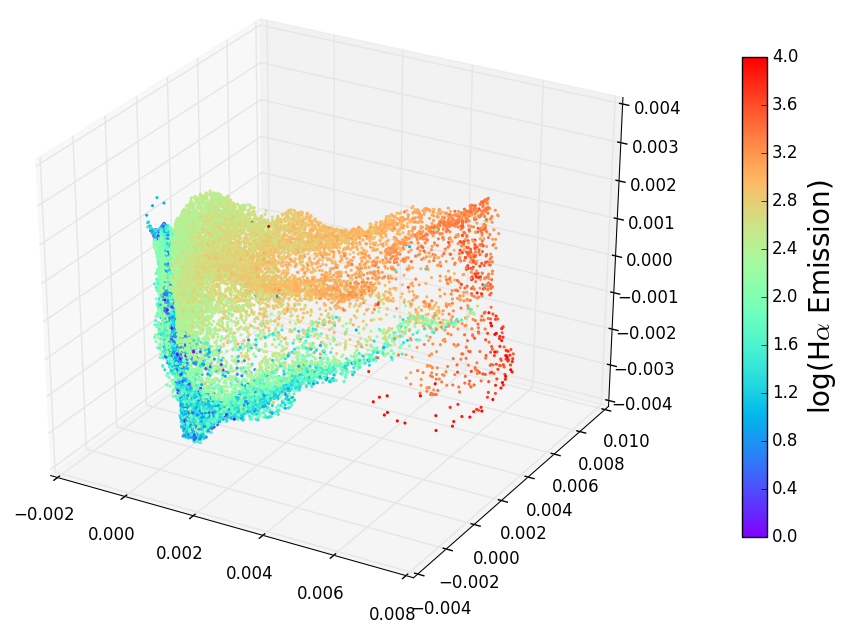}
  \captionof{figure}{\footnotesize
		  {A three-dimensional embedding of the main sample of galaxy spectra
           from the Sloan Digital Sky Survey (approximately 675,000 spectra
           observed in 3750 dimensions). Colors in the above figure indicate
           the strength of Hydrogen alpha emission, a very nonlinear feature
           which requires dozens of dimensions to be captured in a linear embedding
           \citep{Yip2004}. A nonlinear embedding such as the one shown here is
           able to capture that information much more succinctly \citep{Vanderplas2009}. (Figure by O. Grace Telford.)
          }}
  \label{fig:spectra}
\end{minipage}%
\end{figure}

We report run times on two other real world data sets, where the embedding were done solely with \mmani.
The first is the {\tt word2vec} data set\footnote{Downloaded from {\tt GoogleNews-vectors-negative300.bin.gz}.} which contains feature vectors in 300 dimensions for about 3 million words and phrases, extracted from Google News. The vector representation was obtained via a multilayer neural network by \cite{mikolov:13word2vec}.

The second data set contains galaxy spectra from the Sloan Digital Sky Survey\footnote{{\tt www.sdss.org}} \citep{SDSS_DR7}. We extracted a subset of galaxy spectra whose SNR was sufficiently high, known as the {\em main sample}. This set contains fluxes from 675,000 galaxies observed in 3750 spectral bins, preprocessed as described in \cite{TelfordVMcM:16}. Previous manifold learning studies of this data operated on carefully selected subsets of the data in order to circumvent computational difficulties of ML \citep[e.g.][]{Vanderplas2009}. Here for the first time we present a manifold embedding from the {\it entire} sample.

\comment{
The data have been preprocessed by moving them to a common rest-frame wavelength and filling-in missing data using a weighted PCA \jtv{need we say more here? Citation?}.}

\begin{tabular}{lrrrrrr}
&&&\multicolumn{4}{c}{\bf Run time [min]}\\
\cline{4-7}
{\bf Dataset} & {\bf Size $N$} & {\bf Dimensions $D$}
 & {\bf Distances}  & {\bf Embedding} & {\bf R. metric}  & {\bf Total}\\
Spectral & 0.7M & 3750 & 190.5 & 8.9 & 0.1 & 199.5 \\
Word2Vec & 3M & 300 & 107.9 & 44.8 & 0.6 & 153.3\\
\hline

\end{tabular}
\comment{
\begin{figure}
\centering
\includegraphics[scale = 0.9]{word2vec_rmetric_plot.png}
  \captionof{figure}{\footnotesize
		  {3,000,000 words and phrases taken from google news Word2Vec
  			results were embedded into 2 dimensions using spectral
  			embedding and re-scaled. The figure above
  			shows a sample of points displaying the overall shape 
  			of the embedding as well as the estimated push-forward
  			Riemannian metric at a subset of the sampled points.}}
  \label{fig:word2vec_rmetric}
\end{figure}
}%end comment
\section{Conclusion}

% end comment

\mmani{} puts in the hands of scientists and methodologists alike tools
that enable them to apply state of the art manifold learning methods
to data sets of realistic size. The package is easy to use for all
\sklearn~ users, it is extensible and modular. We hope that by
providing this package, non-linear dimension reduction will be benefit
those who most need it: the practitioners exploring large scientific
data sets.

% Acknowledgements should go at the end, before appendices and references

\acks{We would like to acknowledge support for this project
from the National Science Foundation (NSF grant IIS-9988642), the Multidisciplinary Research Program of the Department
of Defense (MURI N00014-00-1-0637), the Department of Defense (62-7760 ``DOD Unclassified Math"), and the Moore/Sloan
Data Science Environment grant.  We are grateful to Grace Telford for creating Figure \ref{fig:spectra}. This project grew from
the Data Science Incubator program\footnote{{\tt http://data.uw.edu/incubator/}} at the University of Washington eScience Institute.}

\vskip 0.2in
\bibliography{mmani-jmlr15}

\end{document}